# A method for the segmentation of images based on thresholding and applied to vesicular textures

Amelia Carolina Sparavigna

Department of Applied Science and Technology, Politecnico di Torino, Torino, Italy

**Abstract:** In image processing, a segmentation is a process of partitioning an image into multiple sets of pixels, that are defined as super-pixels. Each super-pixel is characterized by a label or parameter. Here, we are proposing a method for determining the super-pixels based on the thresholding of the image. This approach is quite useful for studying the images showing vesicular textures.

## 1. Introduction

In image processing, a segmentation is a process of partitioning an image into multiple sets of pixels, defined as super-pixels, in order to have a representation which is simpler than the original one and more useful to the following desired analyses [1]. For this reason, the segmentation of images is often used in many applications of the image processing, and in particular in the medical image processing, where the aim is that of determining the presence of pathologies, and for the stacking of the maps coming from tomography to have the 3D reconstructions [2-4].

The typical use of the image segmentation is that of locate objects, or domains, and boundaries among them. Specifically, the segmentation is a process of assigning a label to every pixel in an image, such that the pixels having the same label share certain characteristics [4]. As a consequence, the result of the segmentation is a set of "segments", or "super-pixels", that are covering the whole image, or a set of contours, that is "edges", extracted from the image. In this case, the segmentation gives the "edge detection".

Several methods exist for segmentation, as we can appreciate from [4]. A discussion of these methods is not the subject of this paper. Here in fact, we want to address this method for applying it to the analysis of the vesicular textures, that we can encounter in geology, materials science and physics. The vesicular textures in images are features that evidence the presence of cavities or holes in the samples; these textures look like those displayed by the volcanic rocks when are pitted with many cavities (known as vesicles) at their surfaces and inside. In such cases, the vesicles are made by gas escaping from cooling lava.

From macrophotography or microscopes (optical, SEM and AFM), or from tomographic devices, we can have 2D maps of pixels representing distributions of vesicular domains. In these cases, it could be interesting to determine the numbers and the size of such domains: this can be obtained by means of segmentation. After some examples illustrating the approach by thresholding to the segmentation, we will consider some studies of the vesicular textures.

## 2. Brightness matrix

Before discussing the segmentation, let us remember something about images and image processing. Let us suppose that our starting point is a RGB source image of $N_x \times N_y$ pixels, reppresented by the three-channel brightness function:

$$b_c : I \to B, I = [1, N_x] \times [1, N_y] \subset \mathsf{N}^2, B = [0,255]^3 \subset \mathsf{N}^3$$

From this function, a grey-tone map can be obtained by giving: $\tilde{\beta}(i,j) = \frac{1}{3}\sum_{c=1}^{3} b_c(i,j)$. Index $c$ corresponds to the three channels R, G, B. The integer indeces $i$ and $j$ are ranging in the $x$ and $y$ directions of the cartesian frame corresponding to the image frame. In this manner, we have, from a RGB image, a brightness map of its grey-tone pixels.

Of course, we could give some statistical parameters of the whole image, such as average brightness $M_0$ and standard deviation $\Sigma_0$:

$$M_0 = \frac{1}{N_x N_y}\sum_{i=1}^{N_x}\sum_{j=1}^{N_y}\tilde{b}(i,j), \quad \Sigma_0^2 = \frac{1}{N_x N_y}\sum_{i=1}^{N_x}\sum_{j=1}^{N_y}\left(\tilde{b}(i,j) - M_o\right)^2$$

From them, many statistical methods of image processing had been developed [5]. The author (ACS) for instance, proposed a statistical approach based on the Coherence Length Diagrams [6-8], for investigating the texture transitions in liquid crystals, evidenced by polarized light microscopy.

### 3. Thresholding and segmentation

It is true that for several physical analysis, the requested measures, which are concerning the domains in the image, are those of simple quantities such as size and area and number of domains. Let us therefore move on with a simple method based on the thresholding of the brightness map. Thresholding is using a clip-level (the threshold value $\tau$), according to which a grey-scale image is turned into a binary image $T$.

$$\tilde{\beta}(i,j) \leq \tau \to T(i,j) = 0 \quad black$$
$$\tilde{\beta}(i,j) > \tau \to T(i,j) = 255 \; white$$

The clip-level has a crucial role in the rendering of the binary image; often it is determined by means of the entropy, evaluated on the histogram of grey-tones of the pixels [9-11]. A simple method is that of using GIMP (GNU Image Manipulation Program), for X Windows systems, and make a visual choice of thresholding.

Once we have the binary image, we have a matrix of pixels containing black and white domains, or segments. Moving from the left/upper corner of this matrix, we focus on black pixels, and characterize each of them by an integer number $k$, acting as a label which will be also the label identifying the segment (super-pixel) to which the pixel belongs. The label is determined according to the labels $k$ of the nearest pixels above (A) and on the left (L) of the considered pixel. If the labels $k_A$ and $k_L$ are the same, their value is the label of the pixel. If the labels are different, the pixel assumes as label the lower value among them. Then, all the pixels having the label with the larger value change their labels into the value of the smaller one. This approach can be easily obtained by logic instructions in any programming language (in the examples here proposed we use Fortran 77). It gives to each of the super-pixels a different label. After this procedure, we have the matrix of labels:

$$T(i,j) = 255 \rightarrow K(i,j) = 0$$
$$T(i,j) = 0 \quad \rightarrow \quad K(i,j) = k, \ k \neq 0, k \subset \mathbb{N}$$

Then, a new map can be proposed, where a grey tone is associated to the label and then to each super-pixel.

Let us see an example based on a detail from an image of a honeycomb structure (Courtesy: Audrius Meskauskas, Wikipedia). In the Figure 1, the steps of segmentation are shown. We start from A, the original image. Then we obtain the grey-tone image B. After, we use GIMP for a simple thresholding to have the black and white image C. After some small adjustments, we obtain the image D, on which we apply the Fortran program for segmentation. In the panel E, we show the super-pixels having different grey tones. Since some are very dark, the panel F shows the output of GIMP Sobel filter for edge detection.

Since each black pixel of the original grey-tone image is labelled by the tone of the super-pixel to which it belongs, we can easily do some calculations. For instance, we can give the distribution of the area (in pixels) of the super-pixels. Since each super-pixel has a different label, we simply count the pixels having the same label. In this manner, we have the areas of the cells of the honeycomb structure. The result is proposed in the Figure 2.

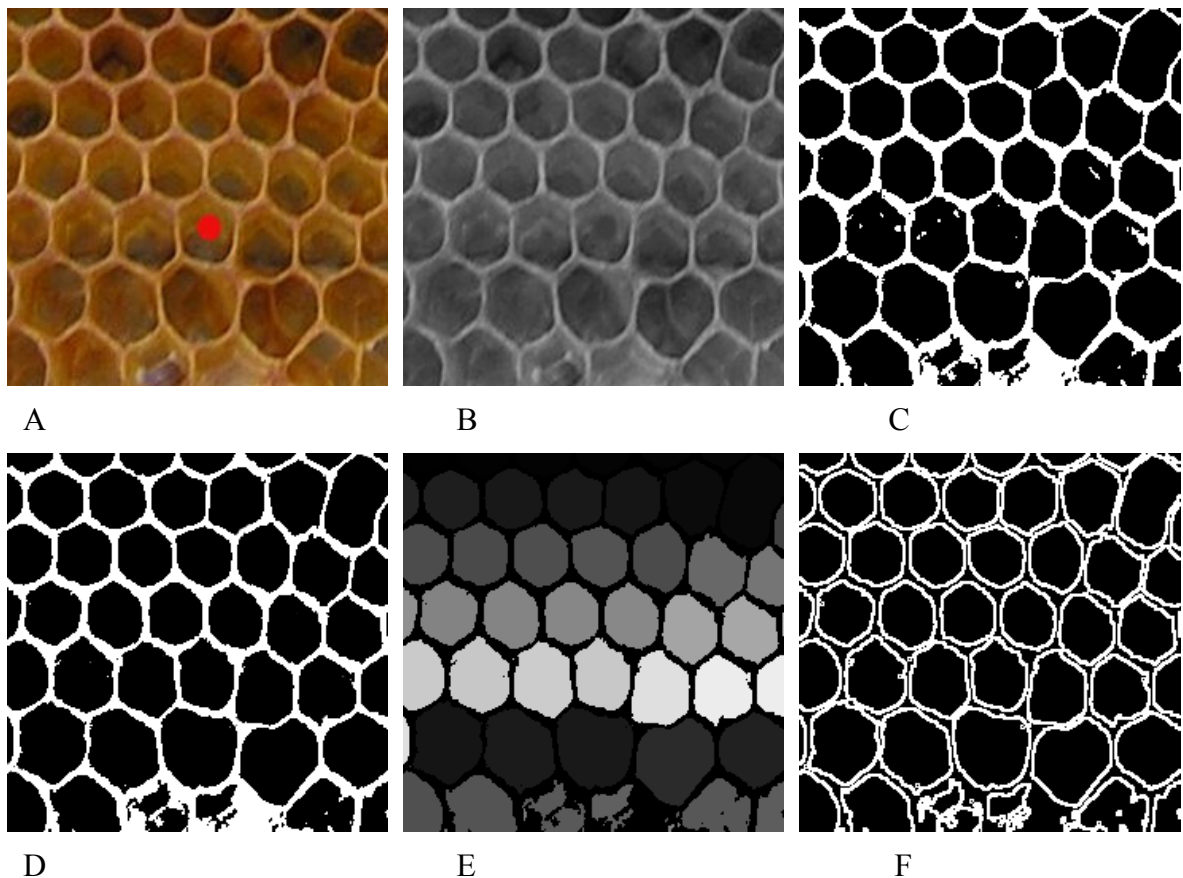

**Figure 1**: A is the original image (240 x 240 pixels). B is the corresponding image in grey tones. C is after thresholding. D is the same as C after some smoothing. E is the result of segmentation according to the Fortran program. The super-pixels are given in grey-tones. Since some of these super-pixels are very dark, they are shown in F by an edge detection (Sobel) obtained using GIMP.

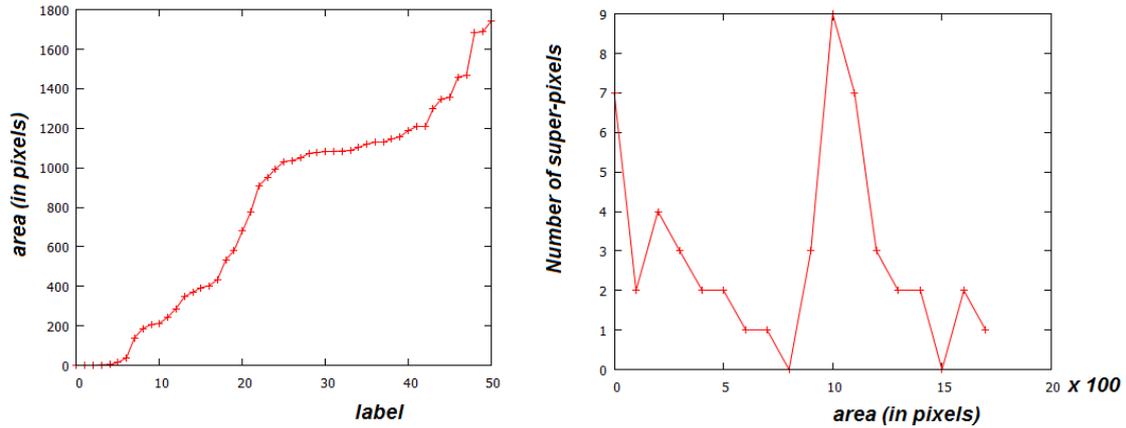

**Figure 2**: Areas of the super-pixels.

Two other examples are given in the Figures 3 and 4. In the Figure 4 we see a detail of an Entobia, a trace fossil in a hard substrate formed by clionaid sponges as a branching network of galleries, often with regular enlargements termed chambers. The detail is from a photograph taken by Mark A. Wilson (Department of Geology, The College of Wooster).

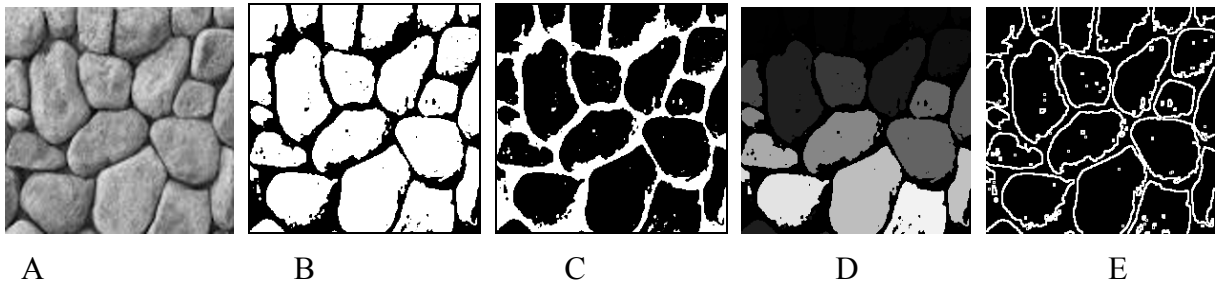

A      B      C      D      E

**Figure 3**: A is the grey-tone image (240 x 240 pixels). B is the image after thresholding. C is B inverted (the inversion of tones is necessary since the program works on black pixels). D is the result of segmentation. The super-pixels are given in grey-tones. Since some of these super-pixels are dark; they are shown by an edge detection, as given in the E panel.

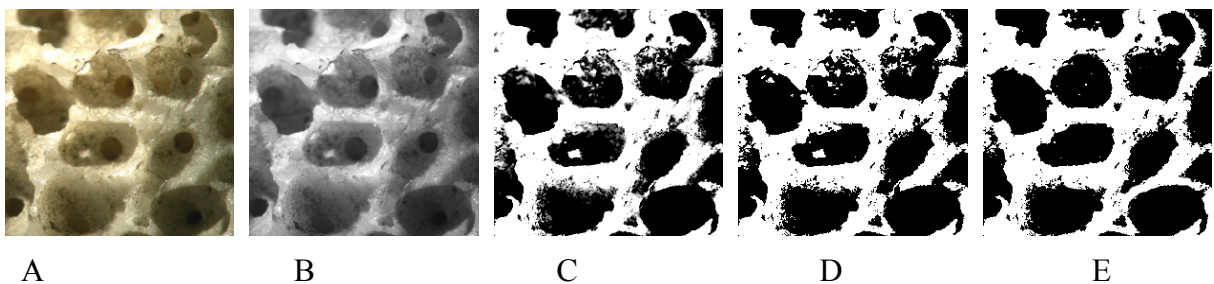

A      B      C      D      E

Figure 4 (continues in the next page)

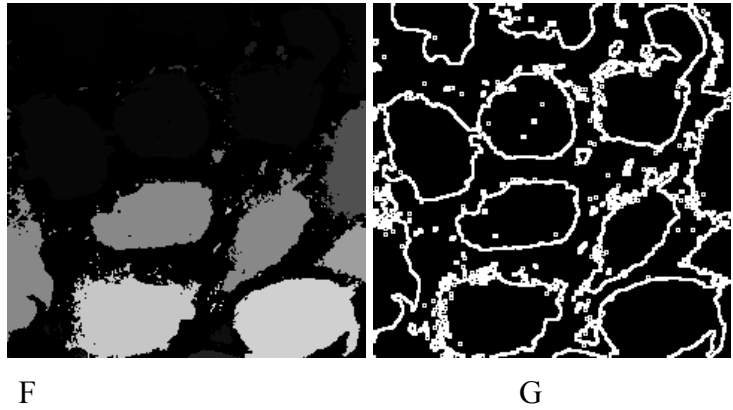

| F | G |

**Figure 4**: A is the original image (240 x 240 pixels). B is its grey-tone image. C is the image after a Retinex filtering provided by GIMP. D is an image after thresholding from C. E is a modified version of D. F is the result of segmentation according to the Fortran program discussed above. The super-pixels are given in grey-tones. Since some of these super-pixels are very dark, again we show them by an edge detection in the G panel.

In the following example (Figure 5), the original image is showing some sponge borings (Entobia made by the genus Cliona). The photograph was taken by Mark A. Wilson (Department of Geology, The College of Wooster). In this case, we show also the areas of the super-pixels, given in the plot of Figure 6.

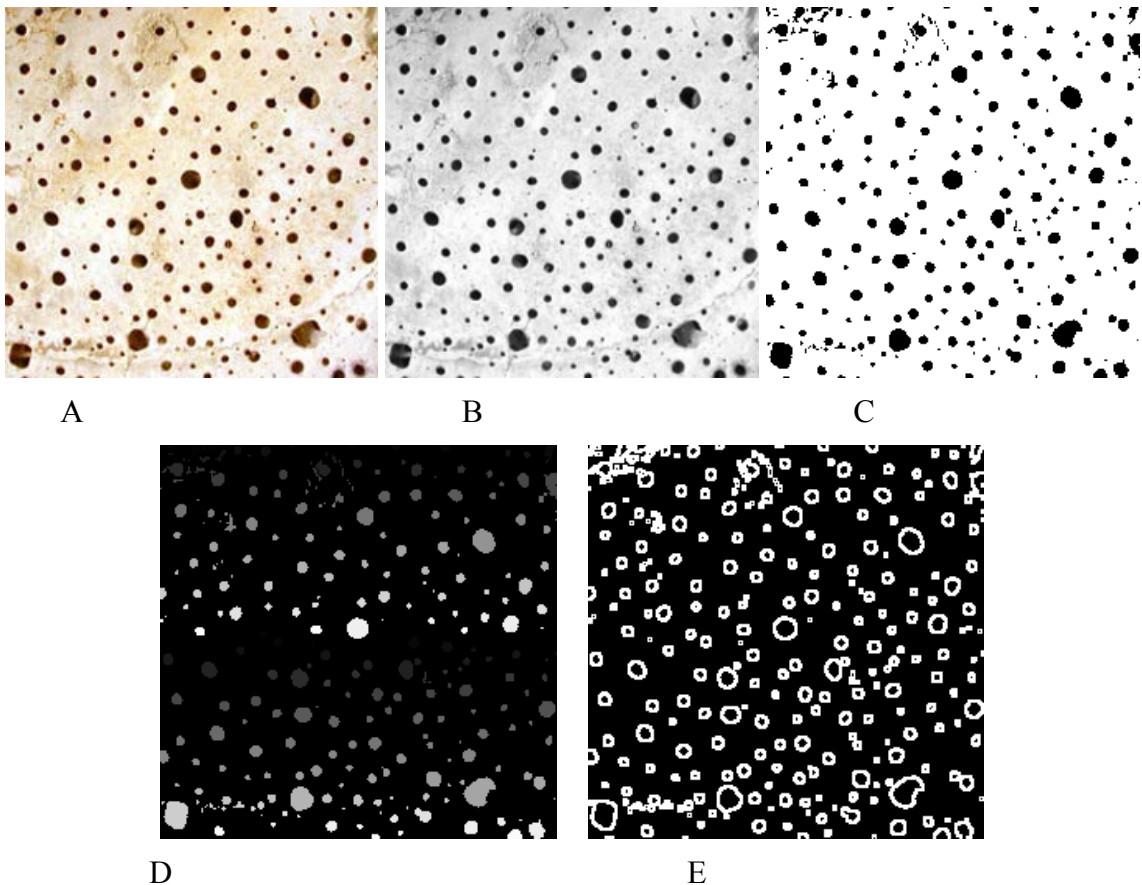

| A | B | C |
| D | E |

**Figure 5**: A is the original image (240 x 240 pixels). B is its grey-tone image. C is after a thresholding. D is the result of segmentation according to the Fortran program discussed above.

The super-pixels are given in grey-tones. Again, all the super-pixels are shown by an edge detection in the E panel.

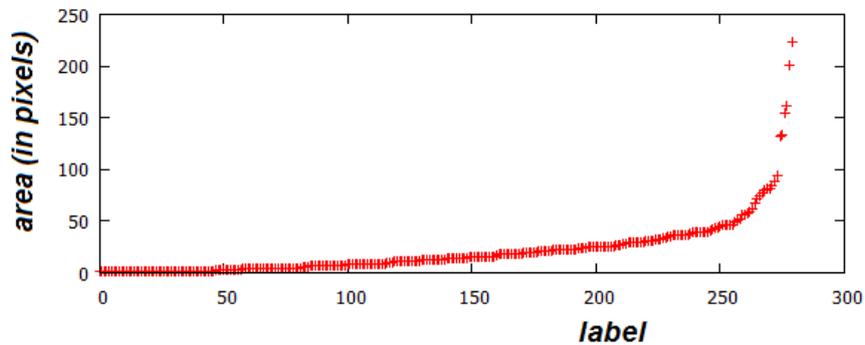

**Figure 6:** Areas of the super-pixels.

**Vesicular textures**

As told previously, the vesicular textures are those containing domains which look like cavities or holes. In fact, in the previous examples, we have already investigated images which are showing cavities, and we have seen that the approach can be very useful to measure the number and the size of the cavities. Let us continue proposing specific examples.

In the Figure 7, the original image is a detail of a photograph taken by Siim Sepp. It is a picture of vesicular basalt with olivine crystals. In the Figure 8, a volcanic sand grain with many vesicles, is viewed with a petrographic microscope (Courtesy: Qfl247, Wikipedia).

Of course other materials, besides the volcanic rocks, have a vesicular structure. In the Figure 9, we see a detail of the surface of a sponge, and in the Figure 10 the surface of a crisp toast.

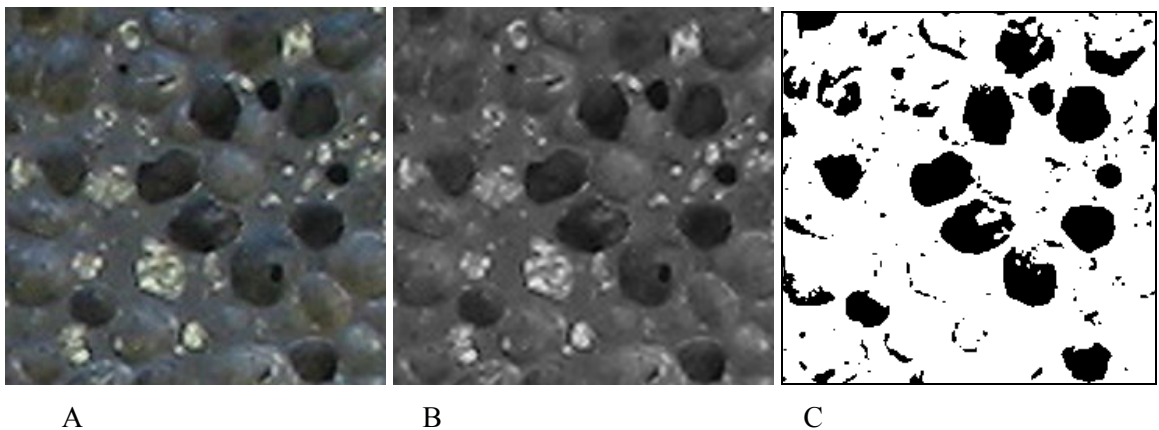

A   B   C

Figure 7 (continues in the next page)

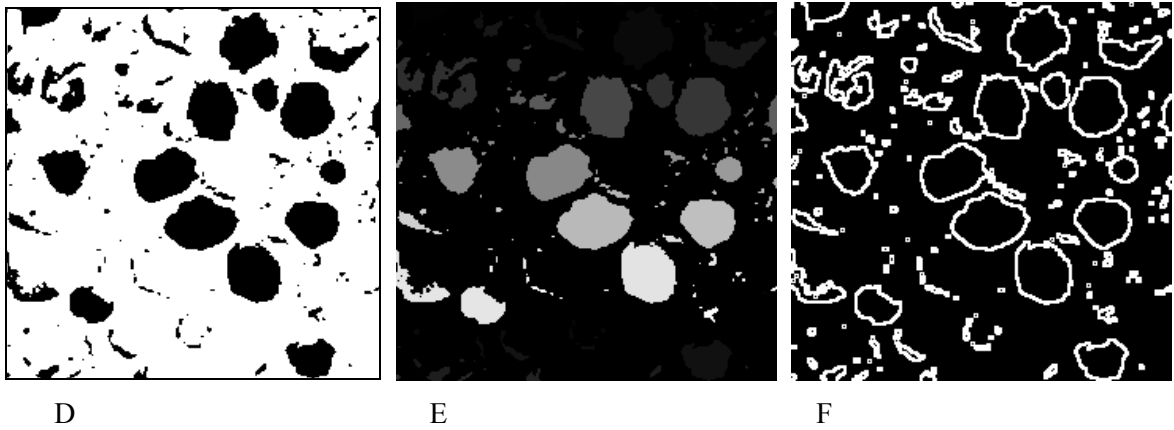

**Figure 7**: A is the original image (240 x 240 pixels). B is the corresponding image in grey tones. C is after thresholding. D is the same as C after some smoothing. E is the result of segmentation, and F the image after edge-detection.

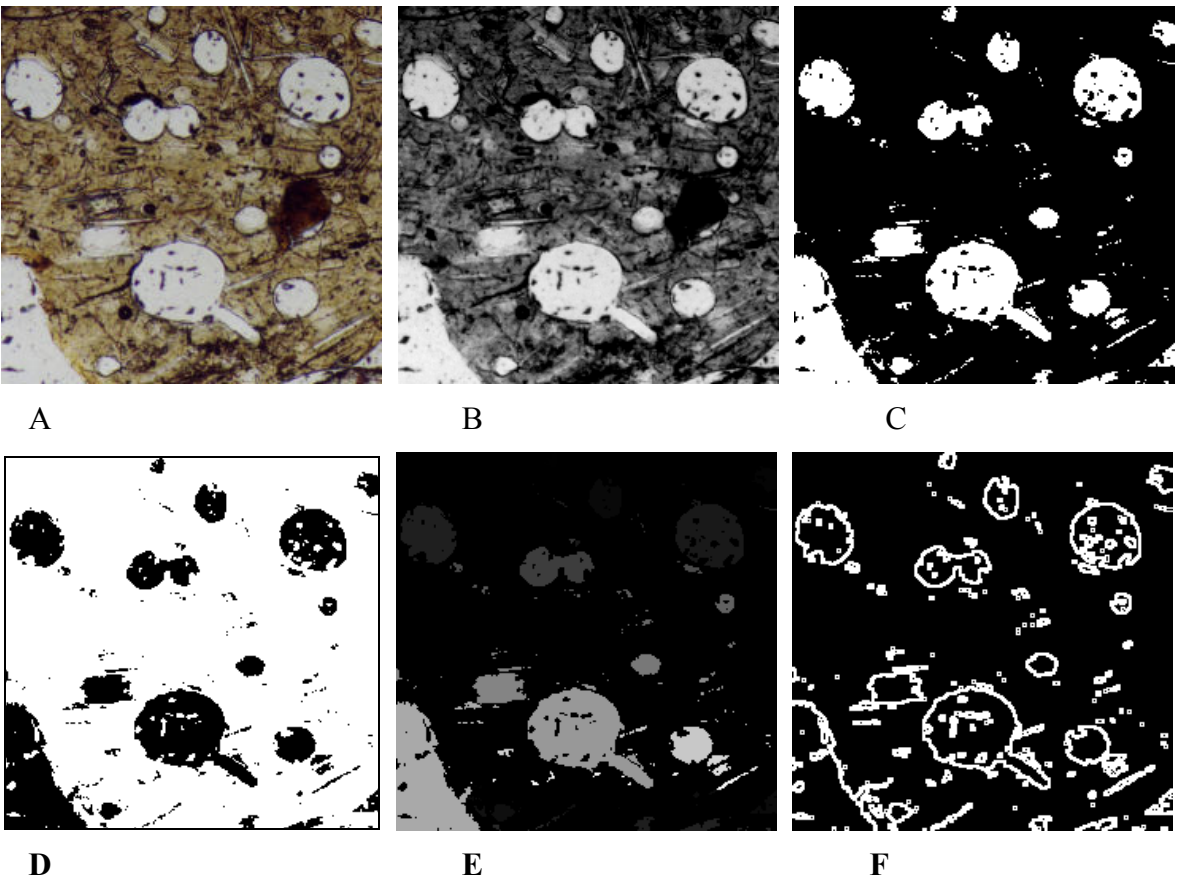

**Figure 8**: A is the original image (240 x 240 pixels). B is the corresponding image in grey tones. C is after thresholding. D is the same as C, but with black and white tones inverted. E is the result of segmentation, and F the image after edge-detection.

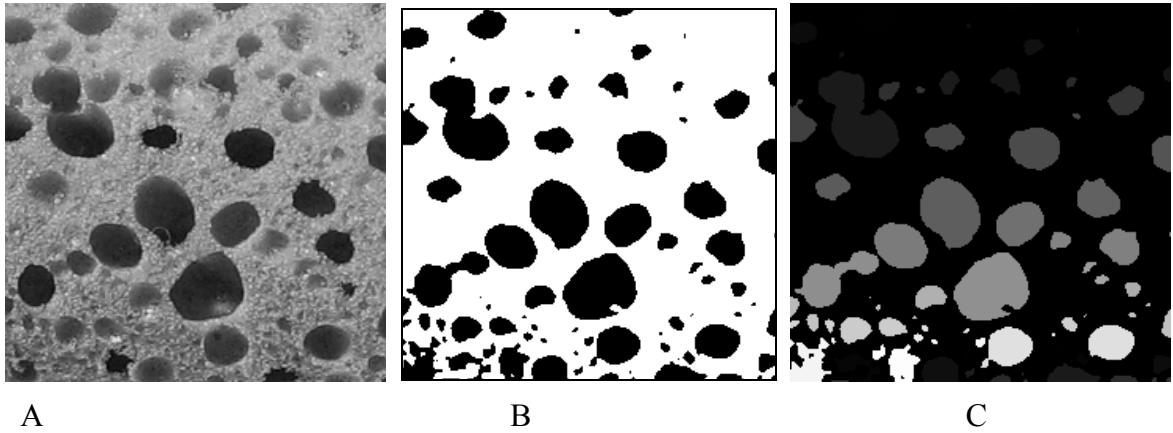

**Figure 9**: A is the grey-tone image (240 x 240 pixels). B is the image after thresholding and C is the result of segmentation. As in the previous cases, we have some super-pixels which are so dark that are not visible in the map.

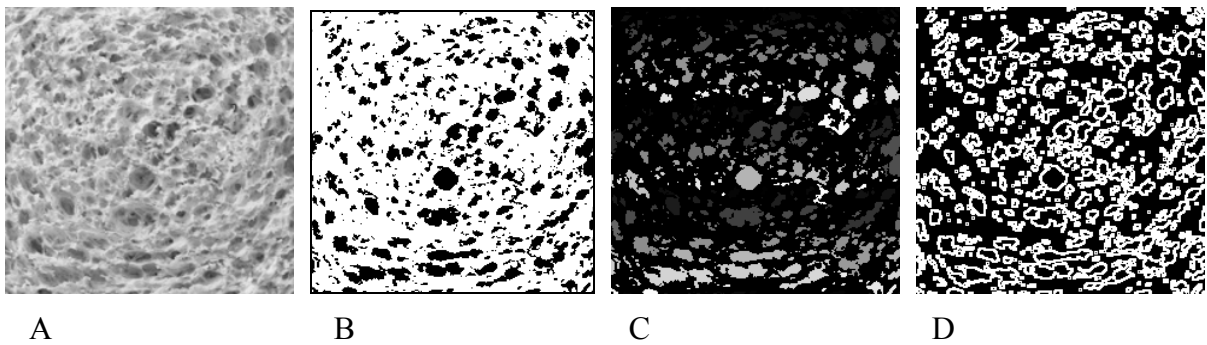

**Figure 10**: A is the grey-tone image (240 x 240 pixels). B is the image after thresholding and C is the result of segmentation. D is the result of edge-detection.

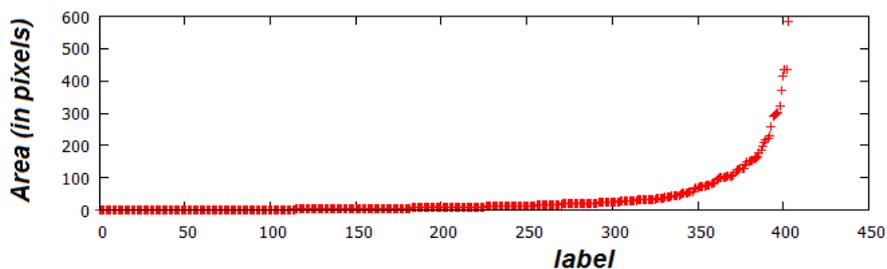

**Figure 11:** Areas of the super-pixels.

As previously told, we can find several cases of materials, having a vesicular structure. The example of the Figure 9, involves the surface of a sponge, and then all such cases can be studied by the proposed segmentation. In the Figure 10, we have used the surface of a crisp toast. Again, all the surfaces of such products and others similar, can be studied. Then, the segmentation we proposed seems being suitable for studying several natural samples and artificial products. The examples given above are coming from photography and macro-photography, but the same approach can be used for images obtained from optical microscopes, SEM and AFM microscopes.